\documentclass[runningheads]{llncs}
\usepackage{graphicx}
\usepackage{tikz}
\usepackage{comment}
\usepackage{amsmath,amssymb} % define this before the line numbering.
\usepackage{algorithmic}
\usepackage{algorithm}
\usepackage{color}
\usepackage{booktabs}
\usepackage[accsupp]{axessibility}  % Improves PDF readability for those with disabilities.

\makeatletter
\newcommand{\printfnsymbol}[1]{%
  \textsuperscript{\@fnsymbol{#1}}%
}
\makeatother

\usepackage[left=22mm,top=22mm]{geometry}

\begin{document}
% \linenumbers
\pagestyle{headings}
\mainmatter
\def\ECCVSubNumber{1137}  % Insert your submission number here

\title{ReFace: Real-time Adversarial Attacks on Face Recognition Systems} % Replace with your title

%\begin{comment}
% \titlerunning{ECCV-22 submission ID \ECCVSubNumber} 
% \authorrunning{ECCV-22 submission ID \ECCVSubNumber} 
% \author{Anonymous ECCV submission}
% \institute{Paper ID \ECCVSubNumber}
%\end{comment}
%******************

%\begin{comment}
\titlerunning{ReFace: Real-time Adversarial Attacks on Face Recognition Systems}
\author{Shehzeen Hussain\thanks{Equal contribution}\inst{1} \and
Todd Huster\printfnsymbol{1}\inst{2} \and
Chris Mesterharm\inst{2} \and
Paarth Neekhara\inst{1} \and
Kevin An\inst{1} \and
Malhar Jere\and
Harshvardhan Sikka\inst{3} \and
Farinaz Koushanfar\inst{1}
}
\authorrunning{S. Hussain et al.}
\institute{UC San Diego, La Jolla, CA 92093, USA \email{\{ssh028,pneekhar,k1an,fkoushanfar\}@eng.ucsd.edu}\\ \and
Peraton Labs, Basking Ridge, NJ 07920, USA
\email{\{thuster,jmesterharm\}@peratonlabs.com}\\ \and
Georgia Institute of Technology, Atlanta, Georgia 30332, USA\\
\email{harshsikka@gatech.edu}}
%\end{comment}
%******************
\maketitle

\begin{abstract}
Deep neural network based face recognition models have been shown to be vulnerable to adversarial examples. However, many of the past attacks require the adversary to solve an input-dependent optimization problem using gradient descent which makes the attack impractical in real-time. These adversarial examples are also tightly coupled to the attacked model and are not as successful in transferring to different models.
In this work, we propose ReFace, a real-time, highly-transferable attack on face recognition models based on Adversarial Transformation Networks (ATNs). ATNs model adversarial example generation as a feed-forward neural network.
% Prior work has used ATNs to generate real-time adversarial examples in the image-classification domain. 
We find that the white-box attack success rate of a pure U-Net ATN falls substantially short of gradient-based attacks like PGD on large face recognition datasets. 
We therefore propose a new architecture for ATNs that closes this gap while maintaining a 10000$\times$ speedup over PGD.
Furthermore, we find that at a given perturbation magnitude, our ATN adversarial perturbations are more effective in transferring to new face recognition models than PGD. 
% We demonstrate that our attacks transfer effectively to models with different architectures, loss functions, and training procedures.
ReFace attacks can successfully deceive commercial face recognition services in a transfer attack setting and reduce face identification accuracy from 82\% to 16.4\% for AWS SearchFaces API and Azure face verification accuracy from 91\% to 50.1\%.

% Recent works have demonstrated that face recognition models are vulnerable to adversarial attacks. However, such attacks require the adversaries to solve an optimization problem for each input they wish to misclassify, making them inapplicable in real-time scenarios. In this work we propose two adversarial perturbation engines to generate real-time adversarial examples for face recognition systems. We first study a perturbation engine that is input-agnostic and adds a single universal perturbation to any input in order to bypass 
% multiple victim face recognition models. Next, we design an input-dependent perturbation engine that relies on an auto-encoder and further improves the attack success rate at the same level of perturbation as the universal perturbation. Finally, we study the extent to which adversarial perturbations transfer across different face recognition systems. We implement and evaluate ReFace to find that it deceives commercial face recognition services from Microsoft Azure Face and Amazon Rekognition in a blackbox setting. We further demonstrate that our attacks are around 100K times faster than existing gradient based attacks against face recognition.

\end{abstract}

\section{Introduction}

Face recognition and verification systems are widely used for identity authentication in government surveillance, military applications, public security settings such as airports, hotels, banks as well as smartphones to unlock applications.
Over recent years, Convolutional Neural Networks (CNNs) have achieved state-of-the-art results on several face recognition and verification benchmarks outperforming traditional computer vision algorithms that rely on hand engineered features. 
With the widespread adoption of face recognition models in surveillance and other security sensitive applications, careful vulnerability analysis is imperative to ensure their safe deployment.

Several works have shown that deep neural networks (DNNs) are vulnerable to adversarial examples, causing the model to make an incorrect prediction with higher confidence~\cite{obfuscated-gradients,Carlini2017TowardsET,goodfellow6572explaining,atscale,limitations}. Particularly, past attacks~\cite{gao2020face,shan2020fawkes} on face recognition systems have garnered immense media attention~\cite{fawkespress,fawkespressnyt} by utilizing projected gradient descent (PGD)~\cite{madry2018towards} based approaches to achieve high fooling success rates. However, designing such adversarial examples requires the adversary to solve an optimization problem for each input. This makes the attack impractical in real-time since the adversary would need to re-solve the data-dependent optimization problem from scratch for every new input.
% Additionally, many of these methods are not feasible against a wide variety of real-world face recognition models due to the unrealistic assumption of full knowledge about victim model architecture, parameters and feature extractors. 

\begin{figure}[htp]
\centering
\includegraphics[width=0.9\textwidth]{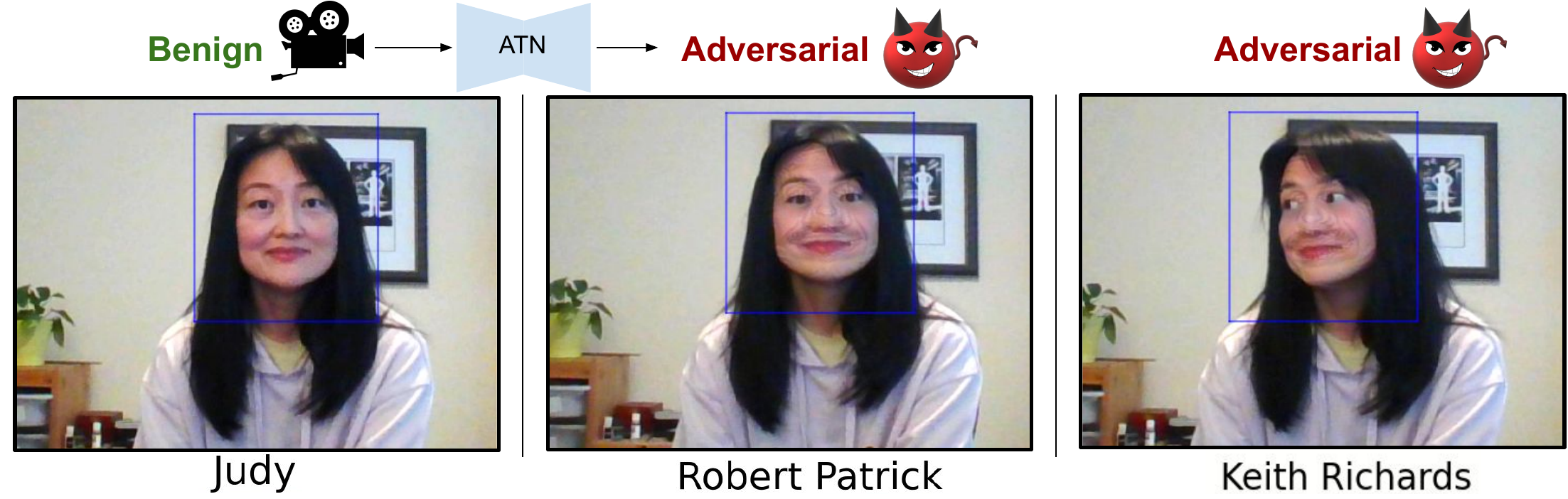}
% \vspace{-0.1in}
\caption{Demonstration of ReFace attack in real-time. Sample screenshots captured from attack demo video posted on our project webpage.}
% \vspace{-0.2in}
\label{fig:demoshot}
\end{figure}

To generate adversarial attacks against classification systems in real-time, some past works, such as Adversarial Transformation Networks (ATNs)~\cite{ATN}, have attempted to learn a perturbation function with a neural network.
ATNs are encoder-decoder neural networks that are trained to generate an adversarial image directly from an input image without having to perform multiple forward-backward passes on the victim classification model during inference, thereby making the attack possible in real time. However, ATNs have only been explored for classification tasks. 
The training objective studied thus far for an ATN is to push the classifier's output outside the decision boundary of the correct class. Unlike a classification model, where model outputs are class probabilities, the output space of a typical face recognition system is an embedding vector. A face recognition system is trained to cluster the embeddings of the same identity together in the embedding space while ensuring they are well separated from the embeddings of other identities. Therefore when attacking such a setup, the attack objective requires the adversary to target the embedding space rather than the decision boundaries of the classifier. 

To perform attacks on face recognition models, we first develop training objectives that target the embedding space of face recognition models and optimize metrics that degrade the identification and verification performance of such models.
To minimize perceptibility of our perturbations, we incorporate Learned Perceptual Image Patch Similarity $L_\textit{pips}$ perceptual loss~\cite{lpipspaper} in addition to the $L_\infty$ constraint during training. Next, to perform real-time attacks, we design a new ATN based on the U-net~\cite{unetpaper} architecture, since U-nets have been notably effective in many prior image-to-image translation tasks~\cite{kandel2020phase,pix2pix2017}. We find that while a U-net based ATN can generate real-time adversarial examples, the attack performance falls short as compared to per-image gradient based attacks such as PGD~\cite{madry2018towards} at the same magnitude of adversarial perturbation. 
This is because gradient-based attacks generate highly tailored adversarial examples that are optimized on a single image.
We address the performance gap between ATN and PGD attacks through neural architectural improvements to our ATN model which we describe in Section~\ref{sec:atnarchsearch}.

Having bridged the gap with gradient based attacks on seen victim models, we evaluate the transferability of our adversarial samples to unseen models.
Since ATNs are trained on a diverse set of images, we find that perturbations generated from an ATN are more transferable to unseen architectures as compared to per-input PGD attack, while being much faster to compute.
To further improve our attack transferability, we adapt our ATN training framework to target an ensemble of face recognition models with various backbone architectures. 
Our best ATN attacks on unseen models successfully reduce the performance of face recognition models to the level of random guessing or worse. We present a demo video of our attack in real-time on our project webpage~\footnote{\url{https://refaceattack.github.io/}} with sample images presented in Figure~\ref{fig:demoshot}. 
Finally, we demonstrate our attack effectiveness against cloud-hosted face recognition APIs in a complete black-box setting.
% We explore neural architectures that are effective at learning to generate adversarial images.
% We identify the Residual U-net as a fast and effective architecture for this task. At the same attack threshold, the Residual U-net is nearly as successful as PGD in performing a white-box attack, and the resulting attack is much faster to compute and transfers more effectively to new models. 
% We validate our approach on face verification and identification problems. 
%Having bridged the gap with gradient based attacks on seen victim models, we evaluate the transferability of our adversarial samples to unseen models.
%Since ATNs are trained on a diverse set of images, we find that perturbations generated from an ATN are more transferable to unseen architectures as compared to per-input PGD attack.
%To further improve our attack transferability, we adapt our ATN training framework to target an ensemble of face recognition models with various backbone architectures. 
%Our best ATN attacks on unseen models successfully reduce face verification AUC to less than 0.5—meaning the model performs worse than random guessing. Finally, we evaluate the our attacks on cloud hosted APIs in a complete black-box setting.
The technical contributions of our work are as follows: 

\begin{itemize}

    \item We propose a real-time attack framework to study the robustness of face recognition systems and demonstrate that our ATNs can synthesize adversarial examples several orders of magnitude faster than existing attacks on face recognition systems while achieving comparable attack success metrics as past works. To the best of our knowledge this is the first real-time attack on face recognition systems, in contrast to previous works which perform gradient based attacks or study real-time attack only in the classification domain. 
    
    \item We bridge the performance gap between real-time ATN attacks and PGD attacks by developing a Residual U-net architecture that allows us to effectively increase the capacity of the ATN (Section~\ref{sec:atnarchsearch}). Our ResU-Net ATN approaches PGD performance in white-box attacks and outperforms PGD on black-box transfer attacks.
    
    \item We develop and release a benchmarking library for face recognition models (Section~\ref{sec:testbed})\footnote{Model test bed to be released upon publication}. This allows us to evaluate our attacks on diverse set of architectures and loss functions. This library may be used to develop more robust face recognition models and to provide benchmarks of models' performance in an adversarial setting.
    
    \item We demonstrate the effectiveness of our real-time attacks on commercial face recognition services such as Amazon Face Rekognition and Microsoft Azure Face. Our attacks reduce face identification accuracy from 82\% to 16.4\% for AWS SearchFaces and face verification accuracy from 91\% to 50.1\% for Microsoft Azure.
\end{itemize}

\section{Related Work}

An adversarial example is an input sample which has been perturbed in a way that is intended to cause misclassification by a victim machine learning model~\cite{biggio,intriguing}.
Prior work on attacks have demonstrated that adversarial examples can circumvent state-of-the-art image classification models while remaining indistinguishable from benign images for humans~\cite{Carlini2017TowardsET,goodfellow6572explaining,madry2018towards,Hussain_2021_WACV,ACM_adv_deepfakes,limitations,transferibility,shi2019curls}. However many of these works are gradient based attacks, which cannot be performed in real-time. To address this limitation, 
the authors of UAPs~\cite{universal} demonstrated that there exist universal \textit{input-agnostic} perturbations which when added to any image will cause the image to be misclassified by a victim network. The existence of such perturbations poses a threat to machine learning models in practical settings since the adversary may simply add the same pre-computed perturbation to a new image and cause misclassification in real-time. Also addressing the real-time challenge, the authors of~\cite{ATN} designed Adversarial Transformation Networks (ATNs) that follow an encoder-decoder architecture and output an adversarial perturbation for each input image, without having to compute gradients from the victim classification model during inference~\cite{ATN}. Unlike UAPs, ATNs generate input-specific perturbations. However these ATN attacks are specific to image classification tasks and cannot be directly used to attack face recognition models that use task-specific model and loss functions as opposed to the standard cross-entropy loss used by classifiers.

Studies on generating adversarial examples for face recognition models are relatively fewer in literature as compared to image classification attacks. The authors of~\cite{rajabi2021practicality} attempt to target face ``classification" networks which operate differently from face recognition networks that perform face verification and identification.  
% These include physical adversarial examples in the form of objects such as glasses~\cite{accessorize_to_crime, sharif2019general} and hats~\cite{komkov2019advhat} that can fool models to make wrong prediction on the person wearing the object. 
Prior works such as~\cite{dong2019efficient,LOTS,shan2020fawkes,gao2020face} generate adversarial examples for face recognition systems by optimizing the perturbation for each image using white-box access to a face-recognition model.
% These attacks rely on white-box manipulations of face recognition models and their feature extractors, which is impractical in real-world scenarios. 
One such attack~\cite{gao2020face} demonstrates that it is possible to generate adversarial faces by optimizing in the model embedding space using PGD~\cite{madry2018towards} and CW~\cite{Carlini2017TowardsET} attack, however reports an attack run-time from 6 seconds to 373 seconds per image while using 2 GPUs. Another gradient based attack Lowkey~\cite{cherepanova2021lowkey} generates image-specific adversarial samples for face recognition models and demonstrates their transferability to public cloud provider APIs, however reports an attack run-time of 32 seconds per image. To generate adversarial examples in black-box settings, the authors of~\cite{dong2019efficient} utilize an evolutionary optimization technique, but require at least 1,000 queries to the target face recognition system before a realistic adversarial face can be synthesized. Similarly, the more recently proposed black-box attack by~\cite{byun2022geometrically} on face recognition systems requires at least 1700 queries to generate successful attacks.
The time for generating adversarial examples using these techniques can potentially bottleneck real-time image upload making the attacks impractical for deployment. 
% The timing bottleneck gets even more significant for videos in which we need to generate adversarial examples for several frames per second.
In contrast, we propose a framework to adversarially modify query images in real-time, such that the performance of face recognition models deteriorate significantly in both white-box and transfer based black-box attack settings.

% Some other attacks on face-recognition systems have proposed GAN based adversarial example generation~\cite{}. These works focus on targeted attacks and rely on images of a target identity to perform the attack. 
% In contrast, we propose a perturbation engine to perform untargeted attacks where the goal is to adversarially modify the query images in real-time and hamper the performance of the face recognition model. 

% All these prior works on attacking face recognition systems require the adversaries to solve an optimization problem for each new input by backpropagating through the victim model. 

\section{Methodology}
\label{sec:methodology}

\subsection{Victim Models}
\label{sec:victimmodels}
\begin{figure*}[htp]
\centering
\includegraphics[width=0.9\textwidth]{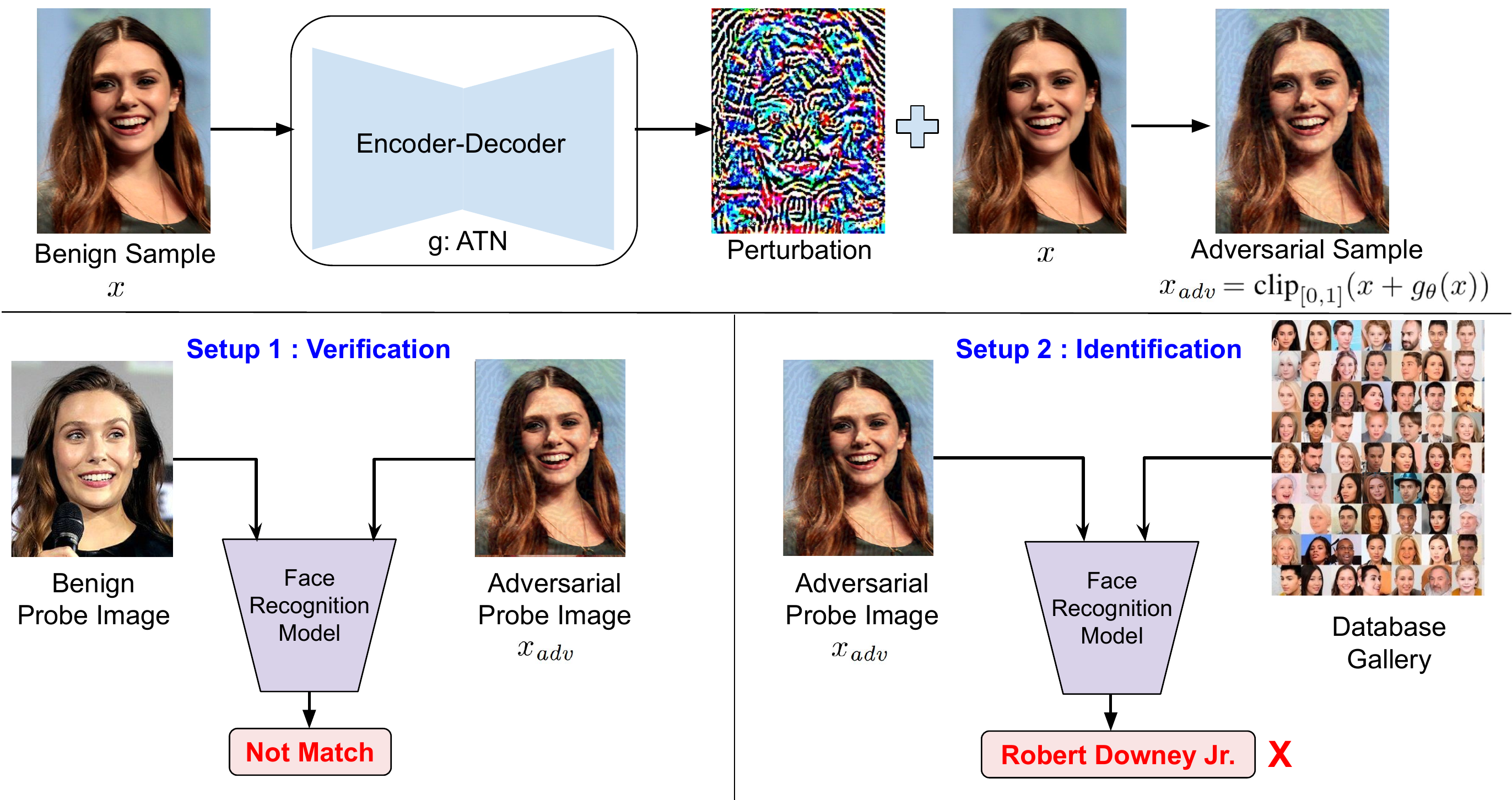}
% \vspace{-0.1in}
\caption{Overview of ReFace adversarial perturbation generator (top) and attack application on face verification and identification systems (bottom). Photo credit: Gage Skidmore}
% \vspace{-0.2in}
\label{fig:generated_adv_samples}
\end{figure*}

A typical face recognition pipeline first detects and crops faces. 
%Each face is then passed through a CNN-based face recognition model that maps the image to an embedding vector. The distances between vectors indicate whether they represent the same person or not. 
%The victim models we consider in our work are CNN based face recognition models. 
Next, they map each cropped image $x$ to an embedding vector $y$ using $ F : x \mapsto \mathit{y} $. 
Typically, such models are trained on a dataset of facial images and identity labels,
with the objective of clustering embeddings of images from the same identity together and ensuring separability between embeddings of images from different identities. State-of-the-art face recognition models are commonly trained with objectives that effectively optimize a cosine distance metric e.g. SphereFace~\cite{Liu2017SphereFaceDH} or DeepID~\cite{Sun2014DeepLF} loss.
% CosFace~\cite{wang2018cosface}, ArcFace~\cite{deng2019arcface} and
During inference, a face recognition model can be used for one of the following goals:

\noindent \textbf{1. Verification} - A face recognition model can be used to verify whether two images belong to the same person or not. In this setting, the model compares the embeddings of two probe images and reports a match if the distance between the embeddings of the two models is below a certain threshold.

\noindent \textbf{2. Identification} - 
In this setting, the face recognition system tries to associate a person with an identity from a set of identities in \textit{gallery images} stored in the system’s database. 
When presented with a \textit{probe image}, the system compares the embedding of the probe image with the gallery images to find the closest matching neighbour in the gallery and determine the identity of the probe image.

In our work, we attack CNN-based face recognition models in real time and assess the success rate against both of the above goals. 
To simplify experimentation, we do not include the detection and cropping step in our attacks pipeline. 
Instead we use the pre-cropped images provided by standard datasets.

\subsection{Problem Formulation}
\label{sec:threatmodel}
We design a perturbation generator that operates in real-time and finds a quasi-imperceptible adversarial perturbation for an input image, which when added to the input causes mis-prediction of the embedding vector thereby degrading the verification and identification performance of the face recognition model. 
When attacking a \textit{face verification} system, we adversarially perturb one of the two probe images. 
In this attack setting, our goal is to reduce the true recall rate of the verification system (performance on positive pairs). 
When attacking a \textit{face identification} system, we assume the probe images have been adversarially perturbed while the dataset of gallery images is benign.
In this attack setting, our goal is to lower the recognition rate of the face identification system. 
\begin{figure}[htp]
        \centering
        \includegraphics[width=0.6\textwidth]{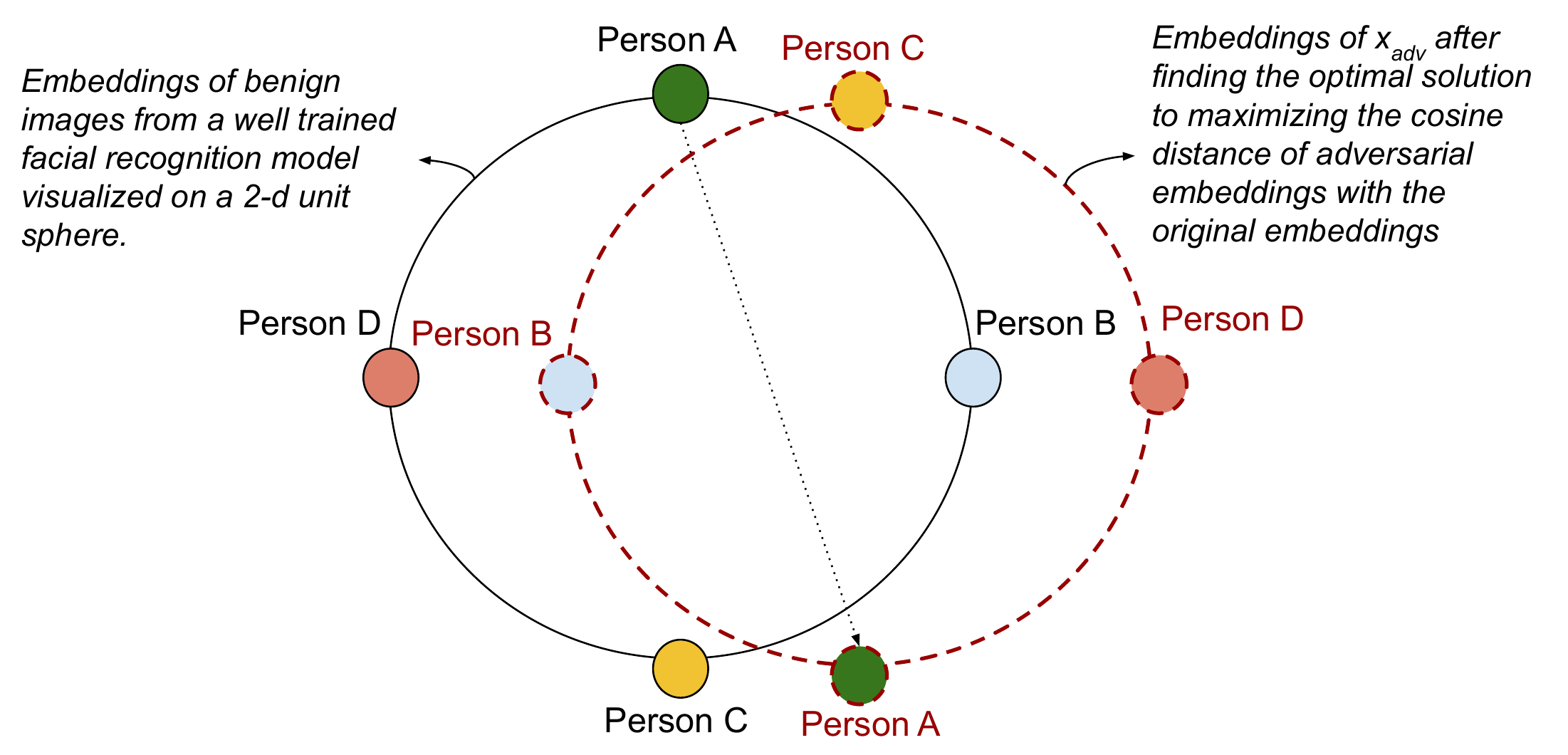}
        % \vspace{-0.1in}
        \caption{\textit{Visualizing the optimum solution to our attack objective:} Our attack objective pushes the originally predicted embedding vectors to the opposite end of the unit sphere thereby hampering the performance of the face-recognition model.}
        \label{fig:attackobjective}
\end{figure}

To achieve the above objectives, we train a perturbation generator $g_\theta$, parameterized by $\theta$, which takes as input an image $x$ and generates an adversarial perturbation $g_\theta(x)$ that can be added to $x$ to synthesize an adversarial example $x_{\textit{adv}}$. 
The optimization objective of $g_\theta$ is to maximize the cosine distance the embeddings of the adversarial and original image, while constraining the amount of the perturbation added to the image.
% \textbf{Proof:} The optimal 
This is different from the objective for fooling classification systems, where the commonly used objective for untargeted attacks is to maximize the cross-entropy loss with the correct label. 
$L_p$ norm is a widely used distance metric for measuring the distortion between the original and adversarial inputs. Prior works~\cite{goodfellow6572explaining} recommend constraining the maximum distortion of any individual pixel using the $L_\infty$ norm. To further reduce the perceptibility of the perturbation we incorporate $L_\textit{pips}$~\cite{lpipspaper} loss during training. $L_\textit{pips}$ distance measures the visual similarity between two images by comparing the embeddings from a pre-trained CNN model. 

Mathematically, our attack objective is as follows:
\begin{eqnarray}
\label{eq:attackobjective}
   & \forall_{x \in X} \textit{ maximize } [d(F(x_{\textit{adv}}), F(x)) - \lambda L_\textit{pips}(x_{\textit{adv}}, x) ]\\
   & \textit{ where } x_{\textit{adv}} = \text{clip}_{[0,1]} (x + g_\theta(x)) \nonumber \\ 
   & \textit{ s.t } ||g_\theta(x)||_\infty < \epsilon \nonumber
\end{eqnarray}
% $$  \textit{ s.t } ||f_\theta(x)||_\infty < \epsilon $$
where $d(F(x_{\textit{adv}}), F(x))$ is the cosine distance between embeddings of the adversarial and original image and $\lambda$ is the loss coeffecient for $L_\textit{pips}$. In Figure~\ref{fig:attackobjective} we illustrate how an optimum solution to the above problem of maximizing the cosine distance completely degrades the performance of a face recognition model. 
% A face recognition model is typically trained with the objective of maximally separating embeddings from different identities while closely clustering embeddings of the same identity on an $n$ dimensional unit sphere. 
A visualization of such embedding clusters for a hypothetical case of four individuals on a 2-D unit sphere is shown on the left in Figure~\ref{fig:attackobjective}. If we were to find the optimum solution to our attack objective in an unbounded attack setting, the embeddings clusters for adversarial images will move to the opposite end of the unit sphere (to maximize the cosine distance). This clearly results in hampering both verification and identification performance of the model since the embeddings of benign and adversarial examples are completely rotated to the opposite ends in the unit sphere.

In our work, we model $g_\theta$ as a neural encoder-decoder architecture called an Adversarial Transformation Network (ATN) (Section~\ref{sec:atnmethod}). %this line is fine
% While the ATN on a victim model with known parameters and architecture, during inference we can use the trained ATN to attack unseen models. We demonstrate in our experiments that adversarial examples generated from the ATN significantly transfer across alternate face recognition models including public APIs such as Microsoft Azure and Amazon Rekognition. 

\subsection{ATN: Adversarial Transformation Network}
\label{sec:atnmethod}
An ATN is a neural network trained to produce adversarial images, with the form $g_{\theta}: \mathcal{X} \to \mathcal{X}$. Since the network only needs one forward pass to compute the perturbation, it is less expensive than an iterative gradient-based optimization procedure. 
We obtain an adversarial image from a benign image using the neural network $\textit{N}_\theta$ as follows: 
\begin{eqnarray}
g_\theta(x) = \epsilon \cdot \tanh (N_\theta(x))
\end{eqnarray}
%g_\theta(x) = \epsilon \cdot \tanh (\texit{N}_\theta(x))

With this formulation we enforce the constraint $||x_\textit{adv} - x||_\infty < \epsilon$ since the output of $\tanh$ is bounded between $[-1, 1]$.

\begin{algorithm}
   \caption{Ensemble attack training procedure}
   \label{alg:algorithmuapatn}
\begin{algorithmic}
    \STATE {\bfseries Inputs:} Victim Models $\mathbb{F} = {F_1, \ldots F_n}$, image dataset $X$
    \STATE {\bfseries Output:} Perturbation engine ($g_\theta$) parameters $\theta$
    \STATE {\bfseries HyperParams:} Learning rate $\alpha$, $L_\infty$ bound $\epsilon$, $L_\textit{pips}$ loss coefficient $\lambda$
    % samples $n$, image dimensionality $N$
   \STATE Initialize ATN: $N_\theta$
   \STATE Batch training images: $X_{\textit{batched}} \gets \textit{Batch}(X)$
   \FOR{$\textit{epoch}$ in $0$ to $N_{epochs}$}
   \FOR{$x$ { in } $X_{\textit{batched}}$}
%   \FOR{$t \text{ in }T$}
        \STATE $x_{\textit{adv}} \gets \text{clip}_{[0,1]} (x + \epsilon \cdot \tanh (N_\theta(x)))$
        \STATE \textit{loss} $\gets 0$
        \FOR{$F_i$ { in } $F$}
            \STATE \textit{loss} $\gets \textit{loss} + ( -d(F_i(x), F_i(x_\textit{adv})) )$
    	\ENDFOR
    	\STATE $\textit{loss} \gets \textit{loss}/\textit{len}(F)$
    	\STATE $\textit{loss} \gets \textit{loss} + \lambda L_\textit{pips}(x_\textit{adv}, x)$
    	\STATE $ \theta \gets \theta - \alpha \cdot \nabla_\theta (\textit{loss}) $
    \ENDFOR
    \ENDFOR
   \STATE {\bfseries return} $\theta$
\end{algorithmic}
\end{algorithm}

We train the ATN to generate adversarial examples using the procedure described in Algorithm~\ref{alg:algorithmuapatn}. Our ATN can be trained to target one or more face recognition models in the model set  $\mathbb{F}$. During each mini-batch iteration, we generate a batch of adversarial images from the ATN and compute the cosine distance between embeddings of benign and adversarial images. We accumulate the loss for all models in the set $\mathbb{F}$ and can optionally add the $L_\textit{pips}$ loss to minimize the perceptibility of the adversarial perturbation. Finally, we backpropogate through all models in the set $\mathbb{F}$ to compute the gradient of the loss with respect to the parameters $\theta$ of the ATN and update the ATN parameters using mini-batch gradient descent with a learning rate $\alpha$. 
Targeting an ensemble of face recognition models during training can result in more transferable adversarial attacks. In our experiments, we verify this hypothesis and demonstrate that ATNs trained to target an ensemble of models result in better transferability to unseen models.

\subsection{The search for an effective ATN architecture}
\label{sec:atnarchsearch}
The input and output domains of the ATN have the same spatial dimension, so a logical choice for the network architecture is a U-net~\cite{unetpaper}. U-nets are commonly used for several image-to-image translation problems. The architecture consists of several down-sampling layers followed by an equal number of up-sampling layers. The feature maps from the down-sampling layers have skip connections that are concatenated to the up-sampling layers with matching resolution. 
Previous work with ATNs used different architectures, but in our preliminary experiments, we found that U-nets were far more effective than alternate architectures at the same level of perturbation. 

However, we still found that there was a large gap between a U-net based ATN and an iterative gradient-based white-box attack, \textit{even on the training data}. This is illustrated in Figure~\ref{fig:pgd2} in our experiments comparing PGD-30 (i.e., 30 iterations of PGD) to the U-net ATN. From the universal approximation theorem~\cite{Pinkus1999ApproximationTO}, a neural network could in principle represent a close approximation of the PGD-30 function. As this neural network would have lower training loss than the U-net ATN, it appears that this architecture is \textit{underfitting}. We therefore explored ways to add capacity to the ATN. We found that adding layers and making the layers wider both led to small gains in performance with diminishing returns. 

\begin{figure}[htp]
         \centering
         \includegraphics[trim={2cm 6.2cm 3cm 5.5cm},clip,width=0.7\columnwidth]{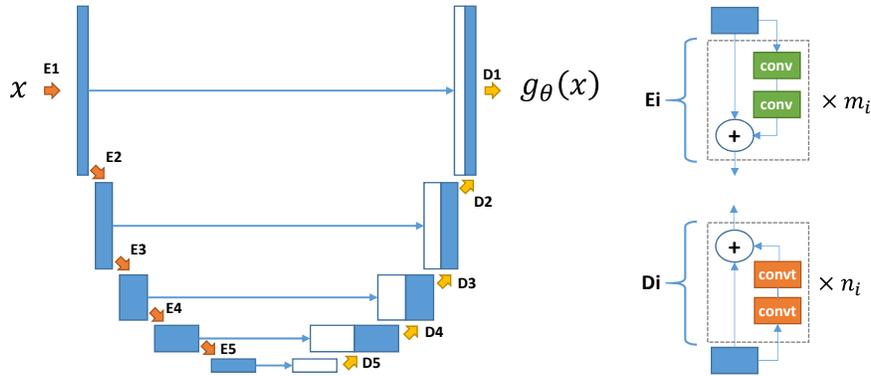}
         % \vspace{-.4in}
         \caption{Residual U-net architecture: We replace the strided convolutions and transposed convolutions in the U-Net architecture with residual blocks. Each residual block contains multiple convolutions (in the encoder) or transposed convolution (in the decoder) layers.}
        %  \vspace{-.4in}
         \label{fig:resunet}
 \end{figure}

One feature of the U-net is that every layer changes the spatial resolution. The deeper layers of the U-net necessarily operate at very low spatial resolutions. Intuitively, it may be useful to be able to express complex hierarchical functions at higher resolutions. We developed a new Residual U-net architecture, illustrated in Figure~\ref{fig:resunet}, that replaces individual convolution and transpose convolution layers in a U-net with groups of residual blocks. 
We use 2-layer pre-activation blocks with ReLU and batch normalization.
One skip connection per downsample is carried over to the decoder, which allows arbitrary numbers of residual blocks at each step. 
We denote the number of blocks in each group as a vectors \textbf{E} and \textbf{D} for the encoder and decoder, respectively. 
While similar architectures have been proposed in the past~\cite{Abdelhafiz2019ResidualDL,Li2018ImprovedBM}, they are not widely used and have not been used in adversarial perturbation literature. 

We found that adding layers in this architecture was considerably more effective than in the pure U-net ATN. 
We performed an architecture search to find an effective balance between computational cost and attack effectiveness. The optimal architecture from this process had five downsampling steps with $\textbf{E} = [1, 1, 2, 3, 5]$ and $\textbf{D} = [1, 1, 1, 1, 1]$. We use a base width of 64 channels and double the width at each downsample step except for the last. 
Using this ResU-net architecture, the ATN approached the performance of PGD-30 (plotted in Figure \ref{fig:pgd2}a)  with roughly 10,000$\times$ less run-time. We refer the readers to the code-base included in our supplementary material for the precise model implementation.

\section{Experiments}
\subsection{Dataset and Models}
\label{sec:testbed}
We developed a benchmarking framework to evaluate both white-box and transfer attack performance of adversarial examples generated using our ATNs. 
% An experiment in this framework defines a set of trained models and sample data available for optimizing an attack. To assess transferability, the attack is evaluated on a disjoint set of data and unseen trained models. 
Our experiments are designed to examine how factors such as network architecture, training loss functions, and random initialization affect the transferability of attacks. 
We used two main CNN architectures for the face recognition models: pre-activation ResNet~\cite{He2016IdentityMI} and Inception-v4~\cite{Szegedy2017Inceptionv4IA}. 
Within these architectures, we varied the number of blocks leading to networks ranging from 22 to 118 layers which were trained with two different loss functions: DeepID~\cite{Sun2014DeepLF} and SphereFace~\cite{Liu2017SphereFaceDH}. 
The face recognition models are trained on the training partition of the VGGFace2 dataset~\cite{Cao2018VGGFace2AD}. 
% Additionally, we report results on the CelebA dataset~\cite{} in our supplementary material. 
We start with the standard crops provided by each dataset and perform random resized cropping for data augmentation during training.

These models are listed in Table~\ref{tab:modeltestbed} and are used to train our ATNs. The model sets include both single and ensemble face recognition models for each architecture to test the effectiveness of ensemble attacks on unseen models. For ensemble models, the reported metrics are averaged over all individual models in the ensemble. 
We split the VGGFace2 validation set into two equal partitions with disjoint identities. We train our ATNs on the first partition and evaluate them on the second. 

% \vspace{-.2in}
\setlength\tabcolsep{3pt}
\begin{table}
\centering
% \resizebox{0.9\columnwidth}{!}{
\begin{tabular}{@{}l|l|cl|cc|c@{}}
  \multicolumn{2}{c}{} & \multicolumn{2}{c}{\emph{Networks}} &  \multicolumn{2}{c}{\emph{Verification}} & \multicolumn{1}{c}{\emph{Identification}} 
  \\ \midrule
  Name & Architecture &  \# Models & Loss & V-AUC & V-Acc. & R1-Acc.
\\ \midrule
RN-SF-1  & ResNet & 1 & SphereFace & 0.99 & 95.2 & 84.4 \\
RN-DID-1  & ResNet & 1 & DeepID & 0.98 & 93.3  & 78.0 \\
IN-SF-1  & InceptionNet & 1 & SphereFace & 0.99 & 94.4 & 78.5 \\
\midrule
RN-SF-6  & ResNet & 6 & SphereFace & 0.99 & 94.3 & 82.0 \\
RN-DID-6   & ResNet & 6 & DeepID & 0.98 & 93.0 &  77.4 \\
IN-SF-4  & InceptionNet & 4 & SphereFace & 0.99 & 94.4 & 78.9 \\
\bottomrule
\end{tabular}
% }
\caption{Victim model sets used for conducting our attack evaluations. Experiments are conducted on both single and ensemble model sets.
The verification and identification metrics are averages over the whole model set reported on the \textit{clean} unperturbed VGGFace2 test set.}
% \vspace{-.6in}
\label{tab:modeltestbed}
\end{table}

\subsection{Evaluation Metrics} 
We evaluate the performance of face recognition models on both verification and identification tasks with the metrics described below.
% Table~\ref{tab:modeltestbed} reports the performance of each model set on clean data.

\noindent \textbf{Face Verification Metrics:}
For each identity in the test set, we prepare all possible pairs of distinct images that have the same identity.  To keep our problem balanced, we randomly sample an equal number of non-matching pairs.  On the test set of VGGFace2, this creates a total of 917,692 verification tests where half have a pair of images with matching identities (positive labels) and half have different identities (negative labels).  Given this binary classification problem, we report the following metrics:

\noindent \textit{1. Verification AUC (V-AUC):} We use the cosine distance between the embedding of the two images along with the verification label to generate a Receiver Operating Characteristic curve (ROC).  Our metric is the standard area under the ROC curve (AUC).

\noindent \textit{2. Verification Accuracy (V-Acc.):} To determine the accuracy, we need a threshold for the cosine distance, across which the example is labelled positive or negative. 
For each model, we set this to \textit{equal error rate} threshold of the model on the (clean) VGGFace2 validation set. 

\noindent \textbf{Face Identification Metric:}
We use the VGGFace2 test set to create a random gallery with 100 unique identities. For each of these 100 identities, we select a probe image with one of the identities appearing in the gallery and compute its distance to each image in the gallery. This creates 100 identification tests. We repeat this gallery test on 1000 random galleries to create a total of 100,000 identification tests. When evaluating attacks, we perturb the probe image and leave the gallery unmodified. We report the \textit{Rank-1 Accuracy (R-1)}, which is the percentage of tests where the image in the gallery with the minimum distance to the probe image has the same identity as the probe image.

%\noindent \textit{1. Rank-1 Accuracy (R-1):} Percentage of tests where the image in the gallery with the minimum distance to the probe image has the same identity as the probe image.
% \vspace{-0.1in}
\subsection{Baseline Attacks}
We compare the effectiveness of ATN attacks against three alternate attacks:
\begin{enumerate}
    \item Universal Adversarial Perturbations (UAP): UAP is a single input-agnostic perturbation vector that can be added to all images to fool the victim models. UAP can be formulated as a simplified ATN where the ATN formulation reduces to : $g_\theta(x) =\epsilon \cdot \tanh(\theta_{h\times w\times c})$. That is, instead of modeling ATN as a neural network, the ATN is modelled using a perturbation vector $\theta_{h\times w\times c}$ which is trained using the same procedure given by Algorithm~\ref{alg:algorithmuapatn}.
    \item Fast Gradient Sign Method (FGSM): FGSM~\cite{goodfellow6572explaining} attack obtains an adversarial example for an image by obtaining the gradient of the optimization objective with respect to the image and then perturbing the image in the direction of the gradient with step size $\epsilon$. That is, $x_\textit{adv} = \textit{clip}_{[0,1]}(x + \epsilon \cdot \textit{sign}(\nabla_xL(x)))$ where $L(x)$ is the optimization objective given by Equation~\ref{eq:attackobjective}.
    \item Projected Gradient Descent (PGD): PGD~\cite{madry2018towards} attack is a multi-step iterative variant of the FGSM attack. Unlike ATNs and UAPs, PGD attack requires several forward and backward passes through a victim face-recognition model to find an adversarial perturbation that is highly optimized for a single image.
    We perform PGD attack as a baseline because it has been commonly adopted by past attacks such as Fawkes~\cite{shan2020fawkes}, Face-Off~\cite{gao2020face} and achieves highest white-box attack success rates. 
    % In our expeirments \textit{PGD-n} refers to PGD with $n$ itertations.
    
\end{enumerate}

\section {Results}
\label{sec:results}
\begin{figure}[htb]
        \centering
        \includegraphics[width=1.0\columnwidth]{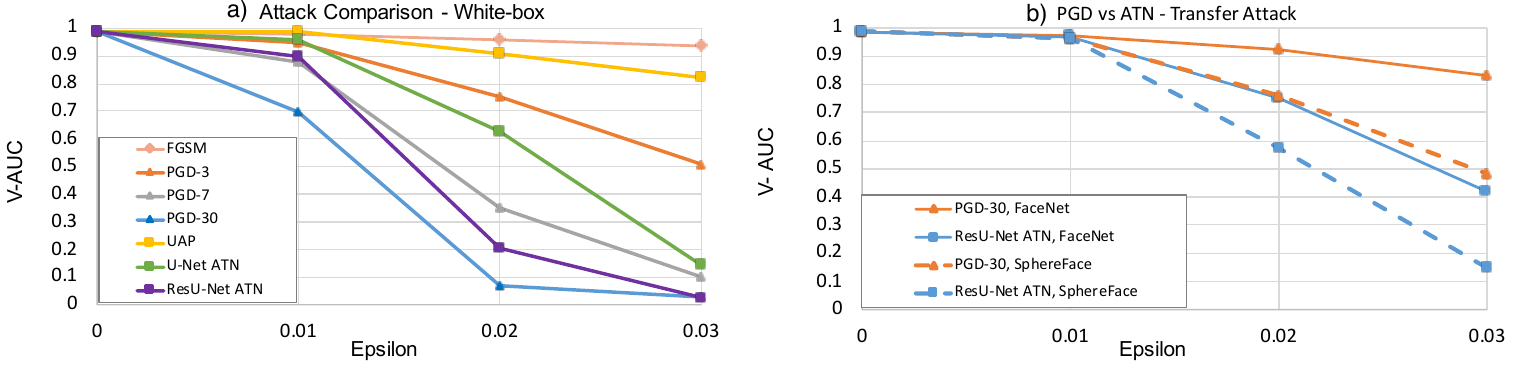}
        \caption{Comparison of PGD and ATN based attacks. (a) compares white-box attacks on the single RN-SF-1 model. (b) compares transfer attacks optimized on the six RN-SF-6 models and evaluated on two different models.}
        \label{fig:pgd2}
\end{figure}

We train six ATN models each targeting one of the model sets listed in Table~\ref{tab:modeltestbed}. The ATNs are trained using mini-batch gradient descent with a batch size $32$ for 500K iterations using Adam optimizer~\cite{Kingma2015AdamAM} with a learning rate $2\textit{e-}4$. 
Our primary evaluation is conducted using the ResU-Net ATN architecture described in Section~\ref{sec:atnarchsearch} at max $L_\infty$ distortion $\epsilon=0.03$ in $[0,1]$ pixel scale.
We present the white-box and transfer attack results of our primary evaluation in Table~\ref{tab:attackresults}. 
Additionally, we present the results and comparisons for the pure U-Net architecture in Section~\ref{sec:pgdvsatn} and comparison against alternate attacks at $\epsilon=[0.01, 0.02, 0.03]$ in Figure~\ref{fig:pgd2}.

% We present examples of adversarial images generated at two different levels $\epsilon=[0.03]$ from each of the techniques in Figure~\ref{fig:exampleimages}. 

% \begin{figure}[h]
%         \centering
%         \includegraphics[width=1.0\columnwidth]{figures/attackresults.pdf}
%         \vspace{-.2in}
%         \caption{}
%         \label{fig:attackresults}
% \end{figure}

\begin{table}[htb]
        \centering
        \includegraphics[width=1.0\columnwidth]{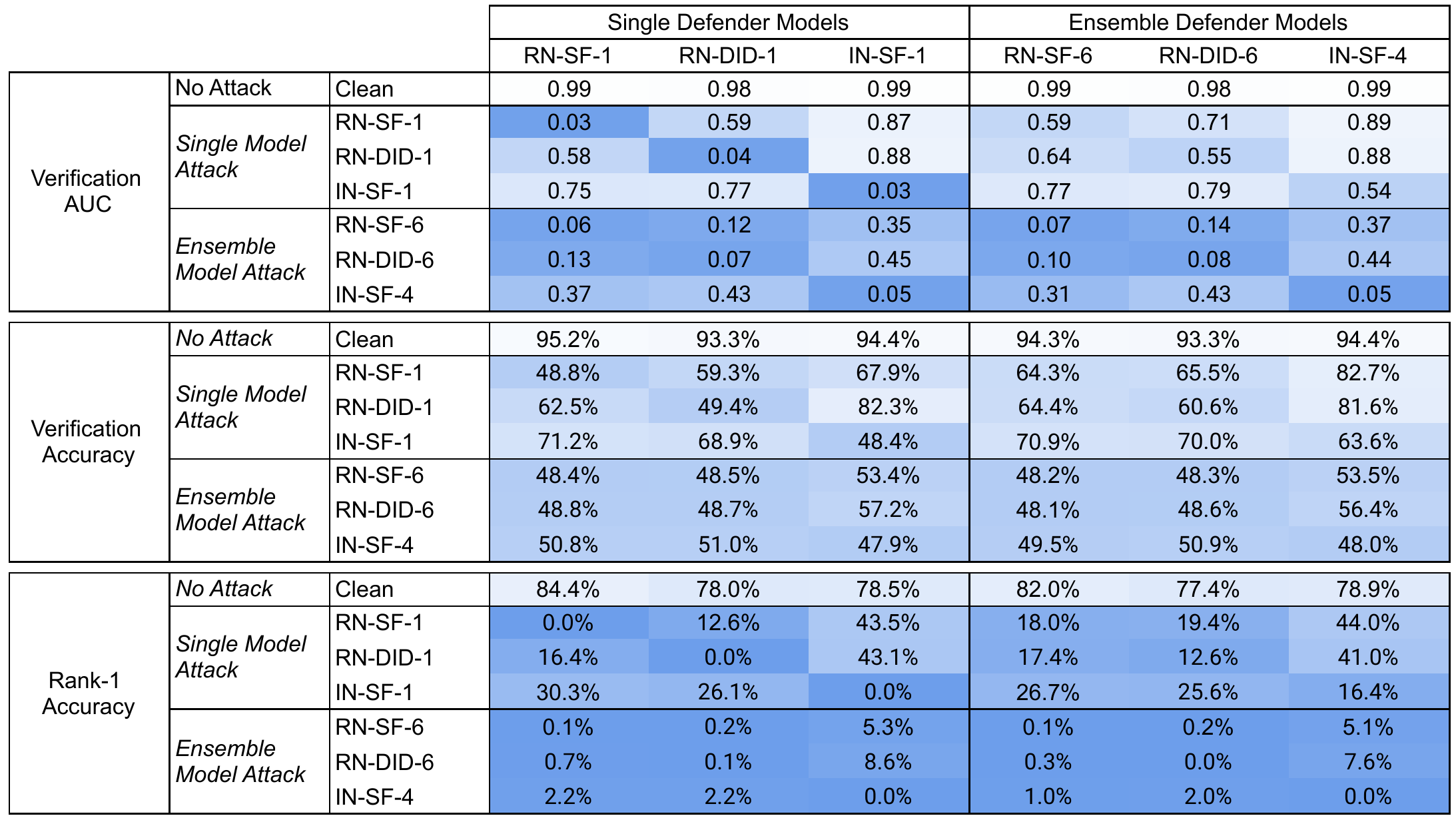}
        \caption{ \footnotesize{White-box and transfer attack results of ATN attack at $\epsilon=0.03$. A lower value for all three metrics indicates a more successful attack. The diagonal entries in each of the three tables represents a white-box attack while all other entries represent a transfer (black-box) attack.  }}
        % \vspace{-0.2in}
        \label{tab:attackresults}
\end{table}

\subsection{Single Model Attack vs Ensemble Attack}
Aversarial perturbations trained on an ensemble of victim models exhibit better transferability across model architectures than those trained on a single model. 
That is, the attack success metrics on unseen models for ATNs trained on ensemble models (RN-SF-6, RN-DID-6 and IN-SF-4) are significantly better than ATNs trained on single models (RN-SF-1, RN-DID-1 and IN-SF-1 respectively). 
The only difference amongst the models in an ensemble is their weight initialization. It is interesting to note that this difference in weight initialization offers enough variance in the model set to train significantly more generalizable perturbations, at the same level of distortion as compared to the single-model attacks.

\begin{figure}[htb]
         \centering
        \includegraphics[width=1.0\columnwidth]{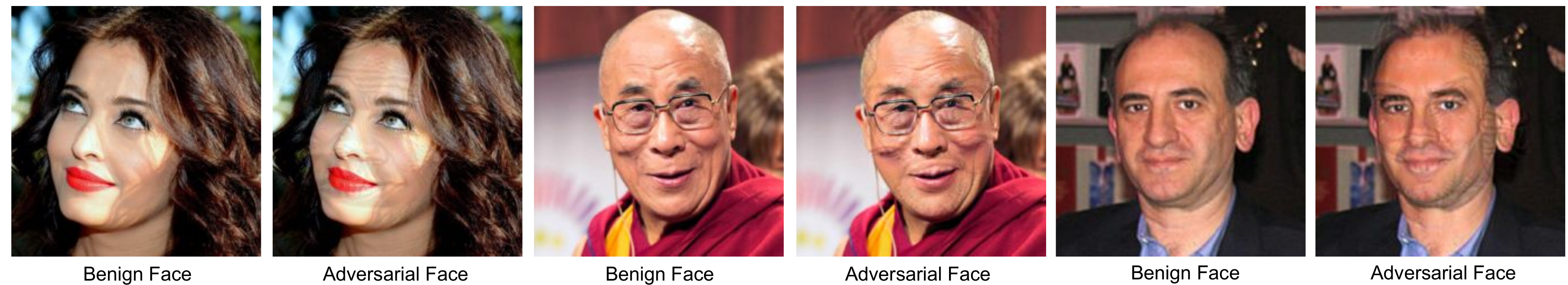}
        % \vspace{-.2in}
         \caption{Sample adversarial images generated at $\epsilon=0.03$ and their benign counterparts. Photo credit: Georges Biard, Christopher Michel, honeyfitz}
        %  \vspace{-.1in}
         \label{fig:sampleimages}
 \end{figure}

\subsection {PGD vs. ATN}
\label{sec:pgdvsatn}
We compare the effectiveness of ATNs and PGD on both seen and unseen models. 
We optimize PGD-30 and ATN attacks on the same surrogate models and perform the attack on a random subset of 10,000 images from the test set. 
For a fair comparison, we drop the $L_\textit{pips}$ term from the loss and train purely to maximize the cosine distance with an $L_{\infty}$ constraint.
Figure~\ref{fig:pgd2}a shows white-box attack success rate of PGD and ATN attacks on the RN-SF-1 model.
As discussed in Section~\ref{sec:atnarchsearch}, Residual U-Net ATN architecture provides a large improvement over a basic U-net and bridges the white-box performance gap between the ATN and PGD-30.

We also performed an ensemble attack on the six models from RN-SF-6.
Running PGD-30 against six surrogate models simultaneously took more than three seconds per image, while the ATN's forward pass was the same complexity as other experiments - more than 10,000$\times$ faster than PGD-30. 
Table~\ref{tab:timing} reports the timing comparison of ATN and PGD attacks. 
In addition to being fast, ATNs learn attacks that generalize effectively to new models. 
We evaluated how well the perturbed images transferred to two different models. 
First, we evaluated against a ResNet+SphereFace model that is similar to the RN-SF-6 models, but has a different number of layers.
Second, we evaluated against an open source model from the FaceNet repository\footnote{https://github.com/davidsandberg/facenet}. 
This model uses a different architecture (Inception ResNet), loss (DeepID) and training procedure. 
We did not do any parameter tuning based on this model, so it serves as an independent validation of the transferability of our attacks.

Figure \ref{fig:pgd2}b compares the attacks at different $L_{\infty}$ thresholds. 
As expected, transferring an attack from RN-SF-6 to the FaceNet model was more difficult than the ResNet+SphereFace model.  
However, in both cases the ATN attack is effective at $\epsilon=0.02$ and transfers much better than PGD to the new models.

\setlength\tabcolsep{6pt}
\begin{table}[htp]
\begin{center}
% \resizebox{0.5\columnwidth}{!}{
\begin{tabular}{l|cc}
\multicolumn{1}{c}{\textit{}}                  & \multicolumn{2}{c}{\textit{Avg Wall-Clock Time (seconds)}} \\ \midrule
Process           & GPU & CPU       \\ \midrule
RN-SF-1    & $2.93\textit{e-}2$ & $1.02\textit{e-}1$\\
\midrule
ATN         &  $2.83\textit{e-}3$  &  $5.67\textit{e-}2$  \\ 
UAP         &  $1.89\textit{e-}4$ & $5.39\textit{e-}3$   \\ 
PGD & $3.73$ & $365.2$ \\
\bottomrule
\end{tabular}
% }
\end{center}
\caption{Average Wall-Clock time in seconds required for generating a single adversarial image on GPU and CPU platforms using different attacks. Time for RN-SF-1 process indicates the forward pass computation time for a single ResNet Face Recognition model.}
\label{tab:timing}
% \vspace{-0.4in}
\end{table}

\subsection {UAP vs ATN}
We find that attacks utilizing ATNs outperform the UAP attacks at the same level of perturbation. Since the goal of finding a single input-agnostic perturbation is more challenging than finding one perturbation per image, a higher amount of distortion is required for a successful attack using UAPs as compared to the ATN based attacks. This is indicated by less successful attack metrics (higher V-AUC) from UAPs in Figure~\ref{fig:pgd2}a.
However, it is important to note that UAPs pose a significant threat to face recognition models since they can be easily shared amongst attackers and are simpler to implement as compared to ATNs.

\subsection{Attacking Public APIs}
\noindent We demonstrate the effectiveness of our attacks against commercial face recognition systems. These systems are black-box, proprietary, and are abstracted away through a web-based API. We evaluate our perturbations against the Amazon (AWS) Rekognition and Microsoft Azure Face services. 

\noindent \textbf{Face Verification:}  In this setting, we target the \textit{CompareFaces} API in AWS and the \textit{verify\_face\_to\_face} API in the Azure Face client. We prepare a total of 1000 image pairs (500 positive and 500 negative) and report the verification metrics in Table~\ref{tab:ATN_verification_accuracies_AWS}.

\noindent \textbf{Face Identification:} We target the \textit{SearchFaces} API in AWS Rekognition. The API accepts a gallery of $N$ faces $x_{1}, x_{2}, x_{3}...x_{N}$ and a query image $x_{q}$, and returns similar faces to the query image from those in the gallery, ranked in order of similarity to the query image. We generate a gallery of $500$ benign faces each with unique identities and $500$ adversarial samples by adversarially perturbing alternate images of the same identities as those in the gallery, resulting in a total of $500$ trials.  We report the Rank-1 accuracy of this experiment in Table~\ref{tab:ATN_verification_accuracies_AWS}. 
% and observe that ATNs are effective at disrupting Rank-1 accuracy of the adversarial probe image among the returned faces.

\begin{table}[htp]
\begin{center}
% \resizebox{0.75\columnwidth}{!}{
\begin{tabular}{l|cc|cc|c}
\multicolumn{1}{c}{} & \multicolumn{4}{c|}{\textit{Verification}} &
\multicolumn{1}{c}{\textit{Identification}}
\\
\multicolumn{1}{c}{} & \multicolumn{2}{c}{{V-Acc. (\%)}} & \multicolumn{2}{c|}{{Recall  (\%)}} & \multicolumn{1}{c}{{Rank-1 Acc.  (\%)}}  \\
 \midrule
Input type                  & AWS     & Azure  & AWS     & Azure & AWS  \\ \midrule
Clean images & 95.5 & 91.0 & 91.0  & 83.0 & 82.0 \\ 
  Ensemble ATN  & 64.7 & 50.1 &  30.2 & 2.1 & 16.4\\ \bottomrule
\end{tabular}
% }
% }
\end{center}
\caption{ATN attack results at $\epsilon=0.03$ against AWS and Azure face recognition APIs. The ATN was trained jointly on RN-SF-6 and IN-SF-4. Recall(\%) indicates the verification accuracy on only the positive pairs in the evaluation set. For verification, we use the default match threshold $0.5$ for both AWS and Azure.}
\label{tab:ATN_verification_accuracies_AWS}
% \vspace{-0.5in}
\end{table}

\section{Conclusion}
We develop real-time attacks using ATNs for fooling face recognition systems. Using our ResU-Net ATN model, we bridge the performance gap between ATN and gradient-based PGD attacks while being several orders of magnitude faster than PGD attacks. 
We demonstrate that adversarial examples generated using ATNs can effectively bypass face recognition systems in both white-box and black-box transfer attack settings. Adversarial examples generated from our framework can bypass commercial face recognition APIs in a complete black-box setting and reduce face identification accuracy from 82\% to 16.4\%. 
% Finally, we demonstrate that adversarial examples generated from our framework can fool unknown face recognition models (including public APIs ) that can be run with minimal computational overhead. 
% We have probed the limits of UAPs in a such a setting and shown that ATNs can extend beyond these limits, both in white box and transfer attack settings.

\section*{Acknowledgements}

This research was funded under Defense Advanced Research Projects Agency contract HR00112090093. This research was, in part, funded by the U.S. Government. The views and conclusions contained in this document are those of the authors and should not be interpreted as representing the official policies, either expressed or implied, of the U.S. Government. Approved for Public Release, Distribution Unlimited.

\bibliographystyle{splncs04}
\bibliography{egbib}
\end{document}